\documentclass{article}

\usepackage[numbers]{natbib}
\usepackage[final]{neurips_2021}
\usepackage[utf8]{inputenc} % allow utf-8 input
\usepackage[T1]{fontenc}    % use 8-bit T1 fonts
\usepackage{hyperref}       % hyperlinks
\usepackage{url}            % simple URL typesetting
\usepackage{booktabs}       % professional-quality tables
\usepackage{amsfonts}       % blackboard math symbols
\usepackage{nicefrac}       % compact symbols for 1/2, etc.
\usepackage{microtype}      % microtypography
\usepackage{xcolor}         % colors

\usepackage{amsfonts,amssymb,multirow,amsmath,arydshln}
\usepackage{graphicx}
\usepackage{picinpar}
\usepackage{wrapfig}
\usepackage{caption}

\title{Weak-shot Fine-grained Classification \\ via Similarity Transfer}

\author{
  Junjie Chen, ~Li Niu\thanks{Corresponding author}, ~Liu Liu, ~Liqing Zhang\footnotemark[1]\\
  MoE Key Lab of Artificial Intelligence,\\
  Department of Computer Science and Engineering,\\
  Shanghai Jiao Tong University\\
  \texttt{\{chen.bys, ustcnewly, shirlley\}@sjtu.edu.cn}, \texttt{zhang-lq@cs.sjtu.edu.cn} \\
}

\begin{document}

\maketitle

\begin{abstract}
Recognizing fine-grained categories remains a challenging task, due to the subtle distinctions among different subordinate categories, which results in the need of abundant annotated samples.
To alleviate the data-hungry problem, we consider the problem of learning novel categories from web data with the support of a clean set of base categories, which is referred to as weak-shot learning.
In this setting, we propose a method called SimTrans to transfer pairwise semantic similarity from base categories to novel categories.
Specifically, we firstly train a similarity net on clean data, and then leverage the transferred similarity to denoise web training data using two simple yet effective strategies.
In addition, we apply adversarial loss on similarity net to enhance the transferability of similarity. 
Comprehensive experiments demonstrate the effectiveness of our weak-shot setting and our SimTrans method. Datasets and codes are available at \href{https://github.com/bcmi/SimTrans-Weak-Shot-Classification}{https://github.com/bcmi/SimTrans-Weak-Shot-Classification}.
\end{abstract}

\section{Introduction}
Deep learning methods have made a significant advance on extensive computer vision tasks.
A large part of this advance has come from the available large-scale labeled datasets.
For fine-grained classification, it is more necessary but more expensive to collect large-scale datasets.
On the one hand, the subtle differences among fine-grained categories dramatically boost the demand for abundant samples.
On the other hand, professional knowledge is usually required to annotate images for enormous subcategories belonging to one category.
As a consequence, fine-grained classification is critically limited by the scarcity of well-labeled training images.

% In practice, we often have a set of base categories with sufficient well-labeled data, and the problem is how to classify novel categories without any well-labeled data, in which base categories and novel categories have no overlap. 
In practice, we often have a set of base categories with sufficient well-labeled data, and the problem is how to learn novel categories with less expense, in which base categories and novel categories have no overlap.
Such problem motivates zero-shot learning \cite{ZSL}, few-shot learning \cite{FSL}, as well as our setting.
To bridge the gap between base (\emph{resp.}, seen) categories and novel (\emph{resp.}, unseen) categories, zero-shot learning requires category-level semantic representation (\emph{e.g.}, word vector \cite{wordvector1} or human annotated attributes \cite{ZSL}) for all categories, while few-shot learning requires a few clean examples (\emph{e.g.}, $5$, $10$) for novel categories.
Despite the great success of zero-shot learning and few-shot learning, they have the following drawbacks:
1) Annotating attributes or a few clean samples require expert knowledge, which is not always available; 
2) Word vector is free, but 
%sometimes ambiguous (\emph{e.g.}, one word has multiple meanings) or unreasonable (\emph{e.g.}, average the word vectors of multiple words in one category name). Besides, word vector is
much weaker than human annotated attributes~\cite{akata2015evaluation}.
Fortunately, large-scale images are freely available from public websites by using category names as queries, which is a promising data source to complement the learning of novel fine-grained categories without any manual annotation.

\begin{figure}[t]
\begin{minipage}[t]{0.49\linewidth}
\centering
\includegraphics[width=1\textwidth]{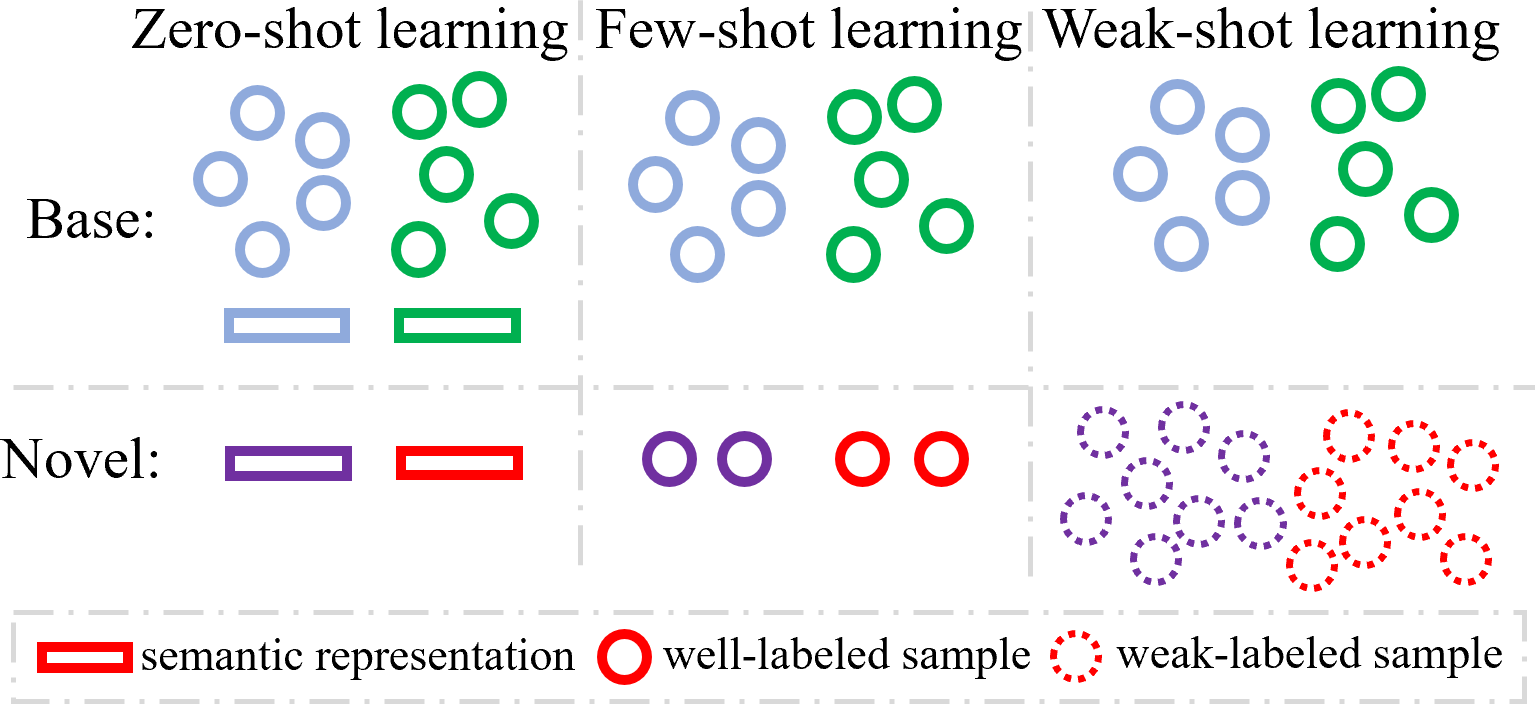}
\caption{  
% Comparison among three learning scenarios.
Comparison among zero-shot learning, few-shot learning, and weak-shot learning.
Different colors indicate different categories.
}
\label{fig:shotlearning}
\end{minipage}\quad
\begin{minipage}[t]{0.499\linewidth}  
\includegraphics[width=1\textwidth]{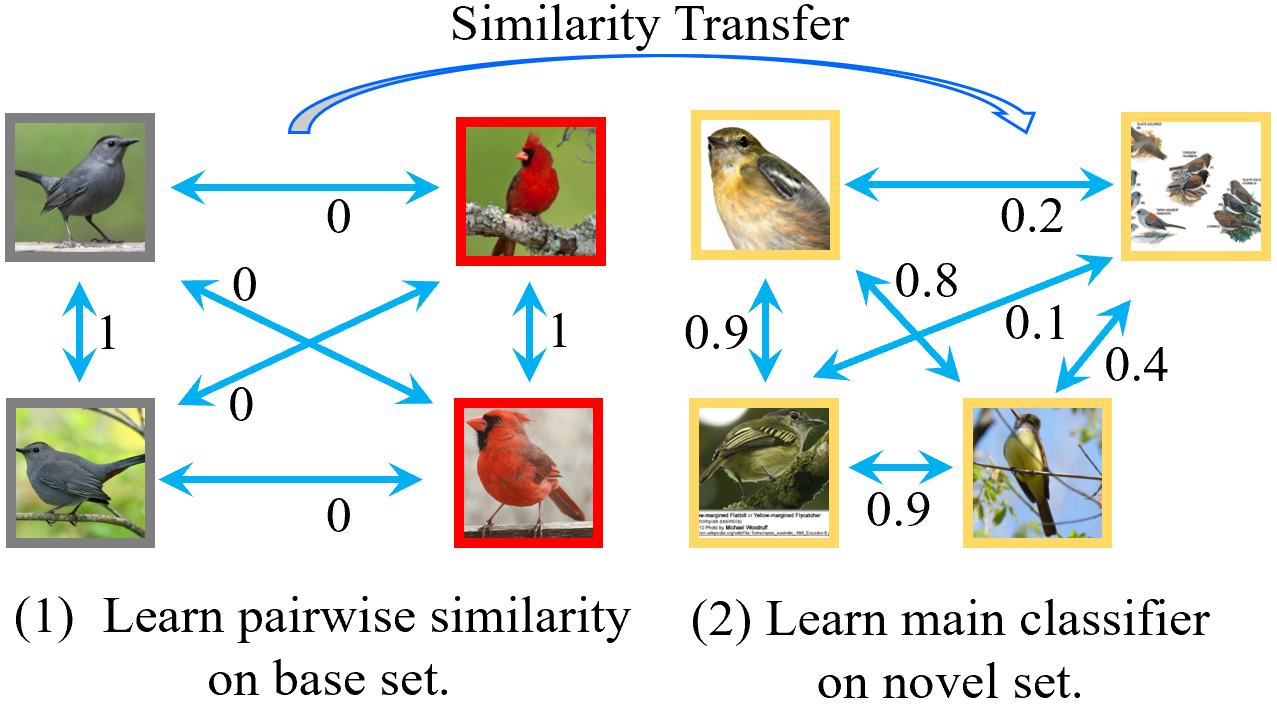}
\caption{  
We transfer the similarity learnt from base training set to enhance the main classifier learnt on novel training set.
The boxes in different colors denote different categories.
The numbers above the arrows indicate similarities. 
}
\label{fig:overview}
\end{minipage}
% \vspace{-15pt}
\end{figure}

Considering the drawbacks of zero/few-shot learning and the accessibility of free web data, we intend to learn novel categories by virtue of web data with the support of a clean set of base categories, which is referred to as weak-shot learning as illustrated in Figure \ref{fig:shotlearning}.
%compared with zero-shot learning and few-shot learning, weak-shot learning does not require any annotation for novel categories.
% \begin{wrapfigure}{r}{0cm}
% \centering
% \includegraphics[width=0.6\textwidth]{figs/shotlearning.png}
% \caption{  
% Comparison among three learning scenarios. Various colors indicate various categories. Different colors indicate different categories.
% }
% \label{fig:shotlearning}
% \centering
% \includegraphics[width=0.6\textwidth]{figs/overview.png}
% \caption{  
% We transfer the similarity learnt from base training set to enhance the main classifier learnt on novel training set.
% The boxes in different colors denote different categories.
% The numbers above the arrows indicate similarities. 
% }
% \label{fig:overview}
% \end{wrapfigure}
Formally, given a set of novel fine-grained categories which are not associated with any clean training images, we collect web images for novel categories as weak-labeled images and meanwhile leverage the clean images from base fine-grained categories.
We refer to the clean (\emph{resp.}, web) image set from base (\emph{resp.}, novel) categories as base (\emph{resp.}, novel) training set.  Weak-shot learning is a useful and important setting. By taking car model classification as an example, we already have the dataset CompCars~\cite{Car} with well-labeled images from base fine-grained categories, but we often need to recognize other fine-grained categories beyond this scope, because there are a huge number of  car models and new car models are also continuously emerging. In this case, we could apply our setting to recognize novel fine-grained categories by collecting the web images for these categories. 
The closest related work to ours is \cite{Meets}, but they further assumed the reliability of word vectors \cite{wordvector1} and the availability of unlabeled test images in the training stage.

In weak-shot setting, the key issue of novel training set is label noise, which will significantly degrade the performance of learnt classifier \cite{Meets,LabelCorrection1}.
We explore using base training set to denoise novel training set, although they have disjoint category sets. 
As illustrated in Figure \ref{fig:overview}, our proposed framework employs the pairwise semantic similarity to bridge the gap between base categories and novel categories.
The pairwise similarity which denotes whether two images belong to the same category is category-agnostic, so it is highly transferable across category sets even if they are disjoint.
Meanwhile, the pairwise similarity can be easily learnt from limited data, indicating that a small set of already annotated images could help learn extensive novel categories (see Section~\ref{sec:sim_transfer}).
Analogously, some methods \cite{MetaBaseline} for few-shot learning transferred similarity from base categories to novel categories, which is directly used for classification.
In contrast, we transfer the pairwise similarity to alleviate the label noise issue of web data.
For learning from web data, some works \cite{CurriculumNet,SelfLearning} also attempted to denoise by similarity.
Nevertheless, their similarities are derived from noisy samples, and likely to be corrupted due to noise overfitting \cite{CleanNet}.
% But their similarities are derived from noisy samples (\emph{e.g.}, feature distances of a model pretrained on web data), and likely to be corrupted due to noise overfitting, leading to sub-optimal results \cite{CleanNet}.

Specifically, our framework consists of two training phases.
Firstly, we train a similarity net (SimNet) \cite{ClusterNet} on base training set, which takes in two images and outputs the semantic similarity. %, \emph{i.e.}, the confidence that they belong to the same category.
Secondly, we apply the trained SimNet to obtain the semantic similarities among web images. In this way, the similarity is transferred from base categories to novel categories. Based on the transferred similarities, we design two simple yet effective methods to assist in learning the main classifier on novel training set.
1) Sample weighting (\emph{i.e.}, assign small weights to the images dissimilar to others) reduces the impact of outliers (web images with incorrect labels) and thus alleviates the problem of noise overfitting.
2) Graph regularization (\emph{i.e.}, pull close the features of semantically similar samples \cite{GCN}) prevents the feature space from being disturbed by noisy labels.
In addition, we propose to apply adversarial loss \cite{GAN} on SimNet to make it indistinguishable for base categories and novel categories, so that the transferability of similarity is enhanced. Since the key of our method is similarity transfer, we name our method \emph{SimTrans}.
We conduct extensive experiments on three fine-grained datasets to demonstrate that the pairwise similarity is highly transferable and dramatically benefits learning from web data, even when the category sets are disjoint. Although the focus of this paper is fine-grained classification, we also explore the effectiveness of our setting and method on a coarse-grained dataset ImageNet \cite{Imagenet}.
We summarize our contributions as
\begin{itemize}
\item We propose to use transferred similarity to denoise web training data in weak-shot classification task, which has never been explored before.
\item We propose two simple yet effective methods using transferred similarity to tackle label noise: sample weighting and graph regularization.
\item One minor contribution is applying adversarial loss to similarity net to enhance the transferability of similarity.
\item Extensive experiments on four datasets demonstrate the practicality of our weak-shot setting and the effectiveness of our SimTrans method.
\end{itemize}

\section{Related Work}
\subsection{Zero-shot and Few-shot Learning}
Zero-shot learning employs category-level semantic representation (\emph{e.g.}, word vector or annotated attributes) to bridge the gap between seen (\emph{resp.}, base) categories and unseen (\emph{resp.}, novel) categories.
A large part of works \cite{ZSL1,ZSL2,ZSL3,ZSL4,gu2020context} learn a mapping between visual features and category-level semantic representations. 
% In addition, some zero-shot learning methods \cite{ZSL_test_1,ZSL_test_2} assume the test images are available during training and employ them to address the domain shift problem.
Our learning scenario is closer to few-shot learning.

Few-shot learning depends on a few clean images (\emph{e.g.}, $5$-shot or $10$-shot) to learn each novel category, which could be roughly categorized as the following three types.
Optimization-based methods \cite{FSL_optim_1,FSL_optim_2} optimize the classifier on a variety of learning tasks (\emph{e.g.}, learn each category by a few images), such that it can solve new learning tasks using only a small number of images (\emph{e.g.}, learn novel categories with a few images).
Memory-based methods \cite{FSL_memory_1,FSL_memory_2,SNAIL} employ memory architectures to store key training images or directly encode fast adaptation algorithms.
Metric-based methods \cite{FSL_metric_1,FSL_metric_2,FSL_metric_3,MetaBaseline} learn a deep representation with a similarity metric in feature space and classify test images in a nearest neighbors manner.

Concerning the problem setting, both zero-shot learning and few-shot learning ignore the freely available web images, whereas we learn novel categories by collecting web images using category names as queries.
Concerning the technical solution, metric-based few-shot learning methods mainly learn image-category similarities and directly recognize test images according to the similarity.
In contrast, we transfer image-image similarities to reveal the semantic relationships among web training images, which are used to denoise the web data for a better classifier.

\subsection{Webly Supervised Learning}
Due to the data-hungry property of deep learning, learning from web data has attracted increasing attention.
Many methods have been proposed to deal with noisy images by outlier removal \cite{OutlierRemovel1,OutlierRemovel2}, robust loss function \cite{RobustLoss1,RobustLoss2,RobustLoss3}, label correction \cite{LabelCorrection1,LabelCorrection2,LabelCorrection3}, multiple instance learning \cite{MIL2,MIL3}, and so on \cite{zhang2020web,AdLoss,yao2019safeguarded,Unreasonable,niu2018learning}.
A prevalent research direction closely related to ours is dealing with label noise using similarities.
To name a few, CurriculumNet \cite{CurriculumNet} computed Euclidean distances between image features, and then designed curriculums according to the distances.
SelfLearning \cite{SelfLearning} employed the cosine similarity between image features to select category prototypes and correct labels.
SOMNet \cite{SOMNet} leveraged self-organizing memory module to construct similarity between image and category, which could simultaneously tackle label noise and background noise of web images.
However, the similarities used in above methods are derived from the noisy training set, which are likely to be corrupted by noise and lead to sub-optimal results \cite{CleanNet,Clothing}.
To alleviate this problem, recent works \cite{CleanNet,Clothing} introduced additional annotations to correct label noise. For example, CleanNet \cite{CleanNet} directly learned the image-category similarity based on verification labels, which involves human verification of noisy images.
Distinctive from the above methods, we do not require any manual annotation on crawled web images.

\subsection{Weak-shot Learning}
In broad sense, weak-shot learning means learning novel categories from cheap weak labels with the support of a set of base categories already having strong labels.
Similar weak-shot setting has been explored in other computer vision tasks, including object detection~\cite{hoffman2014lsda,weakshot-detection1,weakshot-detection2,weakshot-detection3,weakshot-detection4}, semantic segmentation~\cite{weakshot-segmentation2}, and instance segmentation \cite{hu2018learning,weakshot-segmentation1}. 
For weak-shot object detection (also named mixed-supervised or cross-supervised), base categories have bounding box annotations while  novel categories only have image class labels. For weak-shot semantic segmentation, base categories have semantic mask annotations while novel categories only have image class labels.
For weak-shot instance segmentation (also named partially-supervised), base categories have instance mask annotations while novel categories only have bounding box annotations. 
To the best of our knowledge, we are the first work on weak-shot classification, which employs transferred similarity to denoise web training data.

\section{Problem Definition} \label{section:definition}
In the training stage, we have $C_b$ base categories with $N^b_{c}$ clean images for the $c$-th base category.
Our main goal is learning to classify $C_n$ novel categories given $N^n_{c}$ web training images for the $c$-th novel category.
Base categories and novel categories have no overlap.
No extra information (\emph{e.g.}, word vector) is required in the training stage.
In the testing stage, test images come from novel categories. We refer to the above setting as weak-shot learning.

For some real-world applications, test images may come from both base categories and novel categories, which is referred to as generalized weak-shot learning. In this paper, we focus on weak-shot learning, while leaving generalized weak-shot learning to Appendix.

\section{Approach}
Our training stage consists of two training phases: learning similarity net (SimNet) on base training set and learning the main classifier on novel training set.
\subsection{Learning SimNet on Base Training Set} \label{section:simnet}
\begin{wrapfigure}{r}{0cm}
\centering
\includegraphics[width=0.6\textwidth]{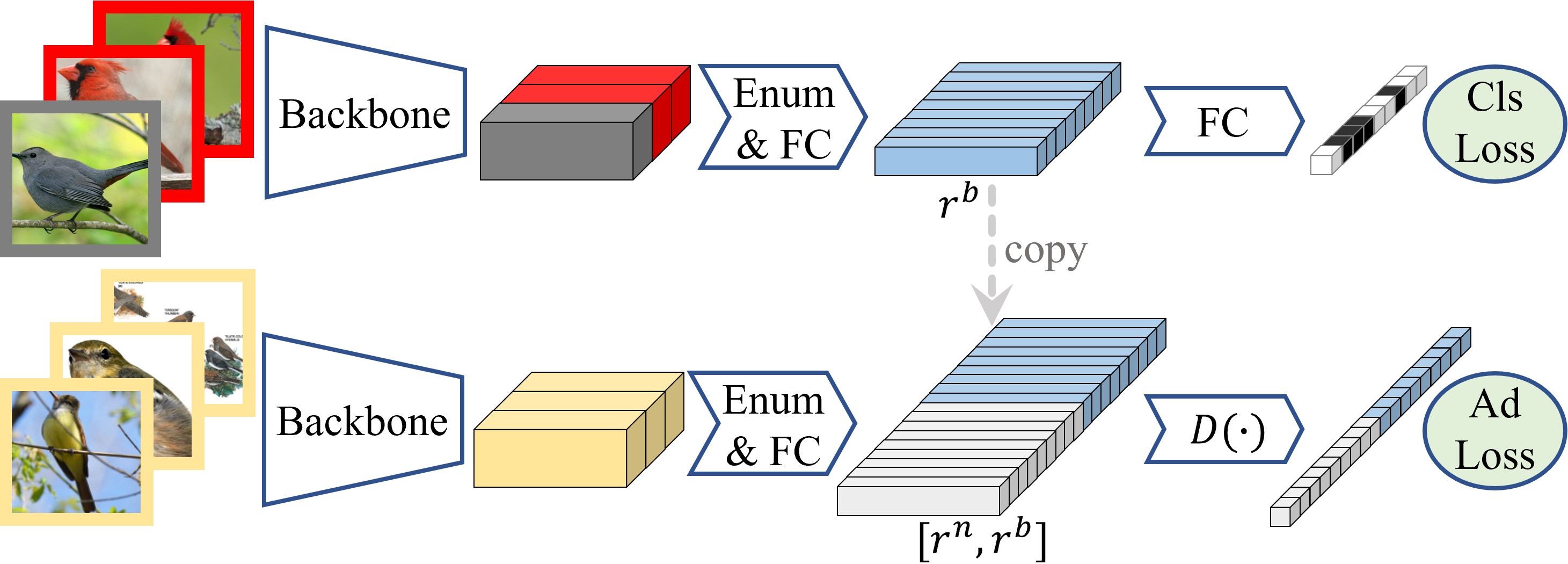}
\caption{ 
Illustration of similarity net with adversarial loss. The top (\emph{resp.}, bottom) pipeline processes base (\emph{resp.}, novel) training set. $\textbf{r}^b$ (\emph{resp.}, $\textbf{r}^n$) represents base (\emph{resp.}, novel) relation features.
}
\label{fig:simnet}
% \vspace{-10pt}
\end{wrapfigure}
Although there are many forms of similarity, we choose an end-to-end deep network SimNet to model the similarity function as in \cite{ClusterNet}.
The architecture of SimNet is shown in the top pipeline of Figure \ref{fig:simnet}, which takes in a mini-batch of images and outputs the pairwise similarity/dissimilarity score for each pair of input images (\emph{e.g.}, $M^2$ pairs for mini-batch size $M$).
The enumeration (Enum) layer simply concatenates each pair of image features which are extracted by the backbone.
For instance, suppose the image features have size $M\times D$ with the feature dimension $D$, then the output of the enumeration layer will have size $M^2\times 2D$.
After that, one fully-connected (FC) layer extracts feature for each concatenated feature pair, which is dubbed as relation feature $\textbf{r}$ of each pair.
Finally, the relation features are supervised by classification loss with binary labels being ``similar" (a pair of images from the same category) and ``dissimilar" (a pair of images from different categories).

Considering the training and testing for SimNet, if we construct the mini-batch randomly, similar and dissimilar pairs will be dramatically imbalanced. Specifically, if there are $C$ categories in a mini-batch, the probability for two images to be from the same category is $\frac{1}{C}$.
To reduce the imbalance between similar pairs and dissimilar pairs, when constructing a mini-batch, we first randomly select $C_m$ ($\ll C$) categories and then randomly select $\frac{M}{C_m}$ images from each selected category, as in \cite{ClusterNet}.
We use $C_m=10$ and $M=100$ for both training and testing of SimNet.

\subsubsection{Adversarial Loss} \label{section:beta}
Ideally, the learnt similarity is category-agnostic, but there may exist domain gap between the relation features of different category sets.
Therefore, we use novel training set in the training stage to further reduce the domain gap between base categories and novel categories, as shown in the bottom pipeline of Figure \ref{fig:simnet}.

Specifically, a discriminator $D(\cdot)$ takes in relation features $\textbf{r}$ in SimNet and recognizes whether they come from base categories ($\textbf{r}^b$) or novel categories ($\textbf{r}^n$).
The SimNet acts as generator, aiming to produce relation features which could not only confuse the discriminator but also benefit the relation classification.
Note that the labels of novel training images are noisy, so we exclude the image pairs from novel categories from the relation classification loss.
Analogous to \cite{GAN}, we optimize SimNet and the discriminator alternatingly in an adversarial manner.
Firstly, we freeze the generator and minimize the adversarial loss of discriminator.
Secondly, with frozen discriminator, we minimize the relation classification loss of SimNet and maximize the adversarial loss of discriminator. 
The classification loss and the adversarial loss are balanced with a hyper-parameter $\beta$, set as $0.1$ via cross-validation.
% The adversarial learning procedure is very trivial and we leave the details to Appendix.
The optimizing is trivial and we leave the details to Appendix.

\subsection{Learning Classifier on Novel Training Set}
\begin{wrapfigure}{r}{0cm}
\centering
\includegraphics[width=0.6\textwidth]{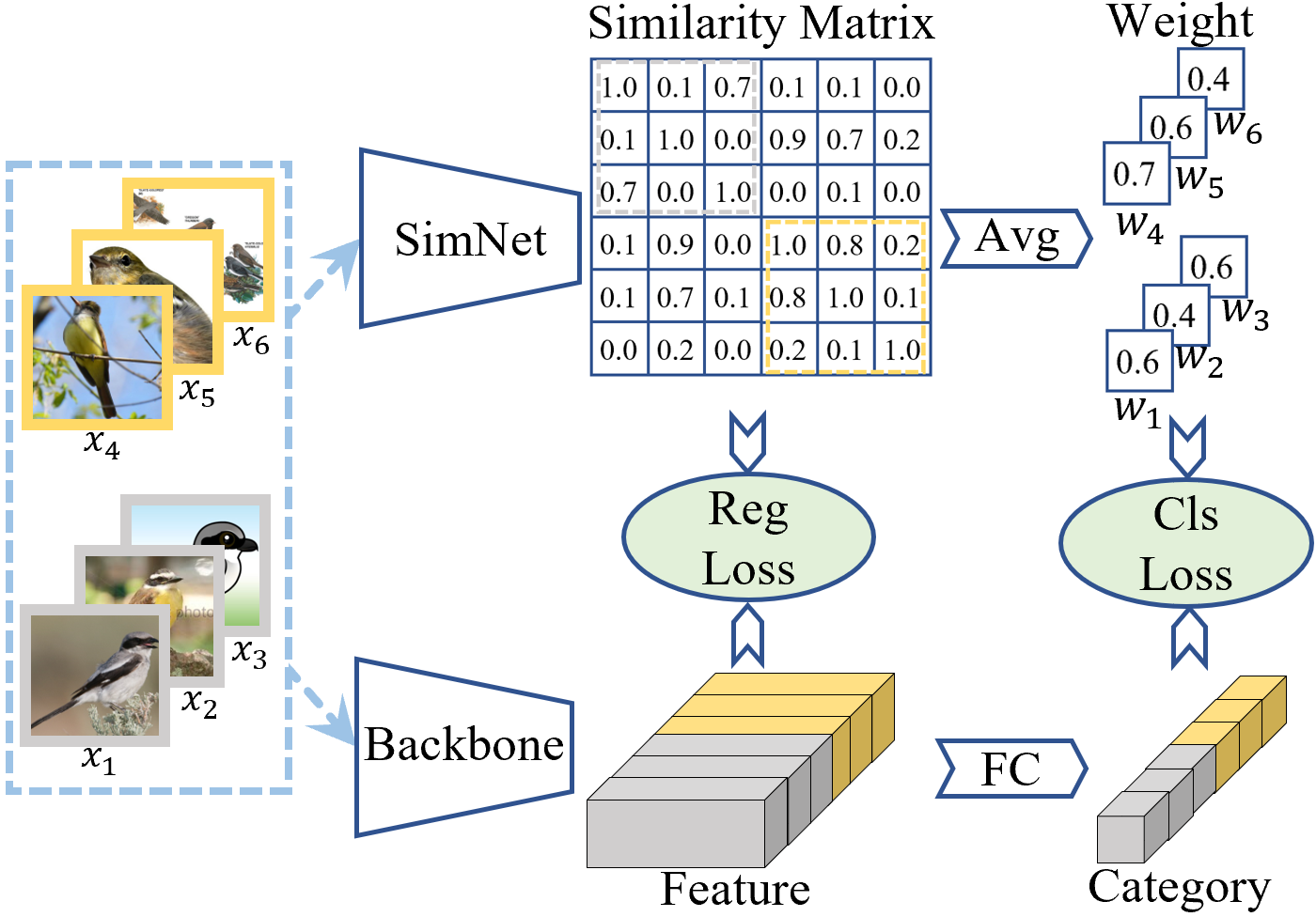}
\caption{ 
The overview of learning novel categories from web data.
The similarity net (SimNet) outputs similarity matrix with pairwise similarities and generates sample weights.
The sample weights are employed to weigh the main classification loss (Cls Loss) of each image, while the similarity matrix is used in graph regularization loss (Reg Loss) based on image features.
}
\label{fig:classifier}
% \vspace{-20pt}
\end{wrapfigure}

Because of the label noise of web images, the performance will be significantly degraded when directly training the classifier using web data.
To address the label noise issue, we employ two simple yet effective methods based on transferred similarities as illustrated in Figure \ref{fig:classifier}. Transferred similarities mean the similarities among novel training samples, which are calculated by SimNet trained on base training set. Next, we will introduce these two methods, \emph{i.e.}, sample weighting and graph regularization, separately.

\subsubsection{Sample Weighting}
For the web images within a novel category, we observe that non-outliers (images with the correct label) are usually dominant, while outliers (images with incorrect labels) are from non-dominant inaccurate categories.
When calculating the semantic similarities among web images within a novel category, outliers are more likely to be dissimilar to most other images.
Therefore, we could determine whether an image is an outlier according to its similarities to other images.
%Simply, we adopt sample weighting to reduce the impact of outliers to classifier.

Formally, for the $c$-th novel category with $N^n_{c}$ web images, we first compute the similarity matrix $\textbf{S}_c \in \mathbb{R}^{N^n_{c}\times N^n_{c}}$, with each entry $s_{c,i,j}$ being pairwise similarity calculated by SimNet pre-trained on base training set.
Although the size of $\textbf{S}_c$ may be large, there are only $N^n_{c}$ times of backbone inference and $N^n_{c}\times N^n_{c}$ times of two FCs inference, which are computationally efficient.
Then, we employ the average of similarities between image $i$ and all the other images as its weight:
\begin{equation}
    w_{c,i}=\frac{1}{N^n_{c}}\sum_{j=1}^{N^n_{c}}{\frac{s_{c,i,j}+s_{c,j,i}}{2}}.
\end{equation}
Then, the sample weights are normalized to have unit mean, \emph{i.e.}, $\bar{w}_{c,i}=\frac{w_{c,i}}{\sum_{j=1}^{N^n_c}{w_{c,j}/N^n_c}}$.
% Based on the analysis before, samples with lower weights $\bar{w}_{c,i}$ are more likely to be outliers.
As analysed before, samples with lower weights $\bar{w}_{c,i}$ are more likely to be outliers.
Finally, we employ weighted classification loss based on sample weights $\bar{w}_{c,i}$. In this way, we assign lower weights to the training losses of outliers, which enables non-outliers to contribute more to learning a robust classifier:
\begin{equation}
    L_{cls\_w}= \frac{1}{M}\sum_{m=1}^{M}{-\bar{w}_{m}{\rm log} f(\textbf{x}_{m})_{y_{m}}},
\end{equation}
where $M$ is the mini-batch size, $\textbf{x}_{m}$ is the $m$-th image in the mini-batch, $\bar{w}_{m}$ is its assigned weight, and $f(\textbf{x}_{m})_{y_m}$ is its classification score corresponding to its category $y_m$.
% \begin{equation}
%     L_{cls\_w}= \frac{1}{M}\sum_{m=1}^{M}{-\bar{w}_{c,m}{\rm log} f(\textbf{x}_{c,m})_{c}},
% \end{equation}
% where $M$ is the mini-batch size, $\textbf{x}_{c,m}$ is the $m$-th image in the mini-batch (belonging to the $c$-th category), and $f(\textbf{x}_{c,m})_{c}$ is its classification score corresponding to category $c$.

\subsubsection{Graph Regularization}
When directly learning on novel training set, the feature graph structure, that is, the similarities among image features, are determined by noisy labels.
In particular, the classification loss implicitly pulls features close for images with the same labels. However, the feature graph structure may be misled by noisy labels, so we attempt to use transferred similarities to rectify the misled feature graph structure.
Specifically, we employ typical graph regularization \cite{GCN} based on transferred similarities to regulate features, which enforces the features of semantically similar images to be close. 

Formally, for each mini-batch of $M$ images, we first compute the similarity matrix $\tilde{\textbf{S}} \in \mathbb{R}^{M\times M}$ using the SimNet pre-trained on base training set.
Regarding the similarity matrix as adjacency matrix, the graph regularization loss is formulated as:
\begin{equation}
    L_{reg}=\sum_{i,j}{\tilde{s}_{i,j}\Vert h(\textbf{x}_i)-h(\textbf{x}_j)\Vert ^2_2},
\end{equation}
where $\tilde{s}_{i,j}$ is each entry in $\tilde{\textbf{S}}$, and $h(\textbf{x}_i)$ is the image feature of $\textbf{x}_i$ extracted by the backbone.

According to \cite{flipnoise}, web images mainly have two types of noise: outlier noise and label-flip noise.
Outlier noise means that an image does not belong to any category within the given category set, and label-flip noise means that an image is erroneously labeled as another category within the given category set.
For the samples with label-flip noise, sample weighting directly discards these samples by assigning lower weights. However, graph regularization can utilize them to maintain reasonable feature graph structure and facilitate feature learning, which could complement sample weighting.

\subsection{The Full Objective} \label{section:alpha}
On the whole, we train the classifier on novel training set by minimizing the weighted classification loss and graph regularization loss:
\begin{equation}
    L_{full}=L_{cls\_w}+\alpha L_{reg},
\end{equation}
where $\alpha$ is a hyper-parameter set as $0.1$ by cross-validation.

\begin{table}[t]
\begin{minipage}[t]{0.5\linewidth}
\caption{
The statistics of splits on three datasets.
\textbf{Category} shows the number of split categories.
\textbf{Train} (\emph{resp.}, \textbf{Test}) indicates the average number of training (\emph{resp.}, test) images for each category.
}
\centering
\begin{tabular}{c|cccc}
\hline \hline
Dataset               & Split & Category & Train & Test \\ \hline
\multirow{2}{*}{Car}  & Base  & 323     & 37      & 34     \\
                      & Novel & 108     & 510     & 36     \\ \hline
\multirow{2}{*}{CUB}  & Base  & 150     & 30      & 30     \\
                      & Novel & 50      & 1000    & 30     \\ \hline
\multirow{2}{*}{FGVC} & Base  & 75      & 67      & 33     \\
                      & Novel & 25      & 1000    & 33     \\ \hline
\end{tabular}
\label{tab:datasets}
\end{minipage}\quad
\begin{minipage}[t]{0.48\linewidth}  
\caption{
Module contributions on CUB dataset.
\textbf{Cls} indicates the cross-entropy classification loss of main classifier.
\textbf{Ad} means training SimNet with adversarial loss.
\textbf{Weight} means sample weighting and \textbf{Reg} means graph regularization.
}
\centering
\begin{tabular}{c|ccc|c}
\hline \hline
Cls & Ad & Weight & Reg & Acc (\%)  \\ \hline
${\surd}$    &             &           &           & 85.4 \\
${\surd}$    &             & ${\surd}$ &           & 90.3 \\
${\surd}$    &             &           & ${\surd}$ & 89.5 \\ \cdashline{1-5}[1.5pt/3pt]
${\surd}$    &             & ${\surd}$ & ${\surd}$ & 91.2 \\
${\surd}$    & ${\surd}$   & ${\surd}$ &           & 90.7 \\
${\surd}$    & ${\surd}$   &           & ${\surd}$ & 90.1 \\ \cdashline{1-5}[1.5pt/3pt]
${\surd}$    & ${\surd}$   & ${\surd}$ & ${\surd}$ & 91.7 \\ \hline
\end{tabular}
\label{tab:ablation}
\end{minipage}
% \vspace{-15pt}
\end{table}

\section{Experiments}
% \vspace{-5pt}
\subsection{Datasets and Implementation}
We conduct experiments based on three fine-grained datasets: CompCars \cite{Car} (Car for short), CUB \cite{CUB}, and FGVC \cite{FGVC}. We split all categories into base categories and novel categories.
The base training/test set and the novel test set are from the original dataset while the novel training set is constructed using web images.
For instance, the novel training set of Car dataset is constructed using the released web images in WebCars \cite{MIL2}, while the novel training sets of CUB and FGVC are constructed by collecting the web images by ourselves.
The statistics of three datasets are summarized in Table \ref{tab:datasets}.
We employ ResNet50 \cite{Resnet} pretrained on ImageNet \cite{Imagenet} as our backbone.
There is no overlap between novel categories and ImageNet $1k$ categories.
Besides, all baselines use the same backbone for a fair comparison.
More details of datasets and implementation are left to Appendix.
%The random seed is set as $0$, and our significant test is left to Appendix.

Note that the focus of this work is fine-grained classification, because 1) it especially requires expert knowledge to annotate labels for fine-grained data (a drawback of few-shot learning); 2) we conjecture that semantic similarity should be more transferable across fine-grained categories.
However, we also explore the effectiveness of our setting and method on a coarse-grained dataset ImageNet \cite{Imagenet}. We leave the experimental results on ImageNet to Appendix.

%We leave the discussion about limitations of the work to Appendix.

\subsection{Ablation Study}
We conduct our ablation study on CUB dataset, considering its prevalence in extensive vision tasks. 
We evaluate the performances of different combinations of our modules, and summarize the results in Table \ref{tab:ablation}.
Solely using sample weighting leads to a dramatic improvement ($90.3\%$ \emph{v.s.} $85.4\%$).
Solely enabling the graph regularization also results in a considerable advance ($89.5\%$ \emph{v.s.} $85.4\%$).
Jointly applying sample weights and graph regularization further improves the performance ($91.2\%$ \emph{v.s.} $85.4\%$).
The adversarial loss on SimNet boosts the performance of classifier for both sample weighting ($90.7\%$ \emph{v.s.} $90.3\%$) and graph regularization ($90.1\%$ \emph{v.s.} $89.5\%$).
Finally, our full method outperforms the baseline by a large margin ($91.7\%$ \emph{v.s.} $85.4\%$).
% Finally, equipped with all modules, our full method outperforms the baseline by a large margin ($91.7\%$ \emph{v.s.} $85.4\%$).

\begin{table}[t]
\begin{minipage}[t]{0.55\linewidth}
\caption{
The first three rows show the performances (\%) of SimNet evaluated by $150$ base categories, $50$ base categories, and $50$ novel categories.
The last row shows the performance (\%) of uniform random guess on $50$ novel categories.
\textbf{PR}, \textbf{RR}, and \textbf{F1} represent precision rate, recall rate, and F1-score, respectively.
}
\centering
\begin{tabular}{c|ccc|ccc}
\hline \hline
\multirow{2}{*}{} & \multicolumn{3}{c|}{\textit{Similar} class} & \multicolumn{3}{c}{\textit{Dissimilar} class} \\
                  & PR       & RR      & F1      & PR        & RR       & F1       \\ \hline
B-150          & 84.4     & 87.7    & 86.0    & 98.6      & 98.2     & 98.4     \\
B-50           & 89.0     & 88.2    & 88.6    & 98.7      & 99.0     & 98.8     \\
N-50          & 88.4     & 87.6    & 88.0    & 98.6      & 98.7     & 98.7     \\ \hline
Rand*           & 10.0     & 50.0    & 16.7    & 90.0      & 50.0     & 64.3     \\ \hline
\end{tabular}
\label{tab:trans}
\end{minipage}\quad
\begin{minipage}[t]{0.4\linewidth}  
\caption{
The performances (\%) of main classifier supported by various similarities learnt from various training sets.
}
\centering
\begin{tabular}{c|c|c}
\hline \hline
Source                          & Type       & Classifier \\ \hline
\multirow{3}{*}{Novel}            & Euclidean & 86.2                \\
                                      & Cosine    & 86.4                \\
                                      & SimNet     & 87.6                \\ \hline
\multirow{3}{*}{\begin{tabular}[c]{@{}c@{}}Novel+\\ Base\end{tabular}} & Euclidean & 87.5                \\
                                      & Cosine    & 87.6                \\
                                      & SimNet     & 89.5                \\ \hline
\multirow{3}{*}{Base}             & Euclidean & 88.8                \\
                                      & Cosine    & 89.1                \\
                                      & SimNet     & 91.7                \\ \hline
\end{tabular}
\label{tab:source_type}
\end{minipage}
% \vspace{-20pt}
\end{table}

\subsection{Investigating Similarity Transfer} \label{sec:sim_transfer}
We evaluate SimNet in various situations to explore the effectiveness of similarity transfer.
Although SimNet is employed to denoise web images, we cannot evaluate its performance on noisy images without ground-truth labels, so we perform evaluation on clean test images.
The performance measurements (\emph{e.g.}, precision rate (PR), recall rate (RR), and F1-score (F1)) are computed at the pair level, instead of image level. 
%Unless otherwise specified, we mainly report the F1-score of \textit{similar} category evaluated on novel test set as the performance of SimNet.
The experiments in this subsection are conducted on CUB dataset.

\subsubsection{The Transferability of Similarity}
The performance gap of SimNet between base categories and novel categories indicates the transferability of learnt similarity.
We first train SimNet on base training set with $150$ categories, and evaluate its performances on base test set with $150$ categories and novel test set with $50$ categories.
We also evaluate on the base test set formed by $50$ random base categories to exclude the influence of different category numbers.
The results are summarized in Table \ref{tab:trans}.
Surprisingly, SimNet trained on base categories achieves a comparable performance for base categories and novel categories (Base-50 \emph{v.s.} Novel-50).
% Note that, we disable adversarial loss here, because it bridges the domain gap between base training images and novel training images, instead of novel test images.
On the whole, the above results suggest that the pairwise similarities are highly transferable across fine-grained categories.

\subsubsection{The Comparison of Similarity Sources and Types}
To demonstrate the superiority of the our transferred similarity, we compare different types of similarities learnt from various training sets (sources) in Table \ref{tab:source_type}.
For \textit{Euclidean} and \textit{Cosine}, pairwise similarity is computed based on the features extracted by a network pretrained on a specific training set (source).
\textit{Cosine} similarity is well-known and we omit the details here.
For \textit{Euclidean}, we adopt the reciprocal of Euclidean distance.
For SimNet, we can train it using different training sets (sources).
By comparing different similarity sources, we observe that all types of similarities learned from base training set perform better, while the performances decrease when novel training set (containing noisy images) is involved.
This verifies that the similarity could be severely corrupted by noisy images, indicating the necessity of learning similarity using clean images even if the category sets are disjoint.
By comparing different similarity types, SimNet performs optimally, showing the superiority of learning similarity in an end-to-end fashion.

\subsubsection{The Impact of the Scale of Base Training Set} \label{sec:base_scale}
\begin{figure}[t]
\begin{center}
\includegraphics[width=1\linewidth]{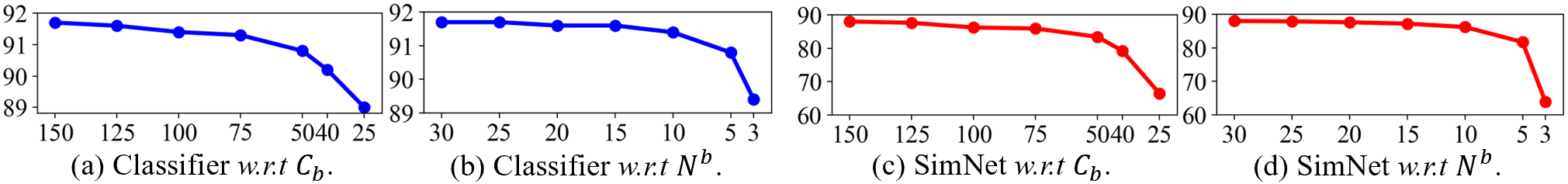}
\end{center}
\vspace{-10pt}
\caption{
Performance variations of main classifier (accuracy \%) or similarity net (F1-score \%) \emph{w.r.t} different numbers of used base categories ($C_b$) or used clean images per base category ($N^b$).
}
\label{fig:limited}
\vspace{-10pt}
\end{figure}

We naturally wonder what is the minimum requirement of the scale of base training set for our similarity net and main classifier to perform well. We explore the scale of base training set from two aspects: category number and image number in each category.
We mainly report the F1-score of \textit{similar} category evaluated on novel test set as the performance of SimNet, and the results are illustrated in Figure \ref{fig:limited}.
The subfigure (a) and (c) shows the robustness when using more than $50$ categories, while the subfigure (b) and (d) shows robustness when using more than $5$ images per category.
Furthermore, we explore the joint impact of two aspects in Table \ref{tab:limited}.
Surprisingly, supported by only $50$ base categories with $5$ images in each category, the main classifier could achieve a satisfactory performance, \emph{i.e.}, $89.3\%$ against $85.4\%$, where $85.4\%$ is the result without the support of base training set.
This study indicates the potential of our setting and method, that is, a small-scale off-the-shelf base training set could facilitate learning extensive novel fine-grained categories.

\subsection{Comparison with Prior Works}
% \subsubsection{Baselines}\label{sec:baselines}

\begin{table}[t]
\begin{minipage}[t]{0.5\linewidth}
\centering
\caption{The performances (\%) of SimNet and main classifier supported by various scales $C\times N$ of base training set, in which ${C}$ denotes the used category number and ${N}$ denotes the used image number in each category.}
\vspace{5pt}
\label{tab:limited}
\begin{tabular}{c|cccc}
\hline \hline
$C\times N$      & 50$\times$5 & 75$\times$5 & 50$\times$10  & 150$\times$30 \\ \hline
SimNet  & 73.2        & 79.1        & 80.6            & 88.0          \\
Classifier & 89.3        & 89.9        & 90.4           & 91.7          \\ \hline
\end{tabular}
\vspace{15pt}
\caption{Accuracies (\%) of representative methods using various numbers of web images on CUB dataset.}
\vspace{5pt}
\begin{tabular}{c|ccccc}
\hline \hline
Number          & 200   & 400 & 600 & 800 & 1000 \\ \hline
Cls Loss        & 78.5  & 82.7    &  84.3    & 85.2     &  85.4    \\
Fusion1         & 80.7  & 84.9    &  87.1    & 88.2     &  88.3    \\
Fusion2         & 81.0  & 85.1    &  87.2    & 88.3     &  88.5    \\ \hline
\textbf{Ours}            & \textbf{84.3}  & \textbf{89.5}    &  \textbf{91.0}    & \textbf{91.6}     &  \textbf{91.7}    \\ \hline
\end{tabular}
\label{tab:webnumber}
% \centering
% \caption{Accuracies (\%) of different methods on three datasets in the generalized weak-shot setting. The best results are highlighted in boldface.
% }
% \begin{tabular}{c|ccc}
% \hline \hline
% Method                & {Car} & {CUB} & {FGVC} \\ \hline
% Cls Loss   & 76.3         & 73.3         & 76.9            \\
% SOMNet+MetaBaseline       &   77.5     &   74.8      &  78.8  \\
% DivideMix+MetaBaseline &  78.1        &    75.0      &  79.3      \\ \hline
% SimilarityTransfer (Ours)     &   \textbf{80.4}     &  \textbf{76.9}     & \textbf{81.7}  \\ \hline
% \end{tabular}
% \label{tab:Gen}
\end{minipage}\quad
\begin{minipage}[t]{0.45\linewidth}  
\caption{
Accuracies (\%) of various methods on three datasets in the weak-shot learning. The best results are highlighted in boldface.
}
\centering
\begin{tabular}{c|ccc}
\hline \hline
Method                   & { Car } & { CUB } & { FGVC } \\ \hline
Cls Loss             & 83.1            & 85.4            & 86.6             \\
SelfLearning                 & 85.1            & 87.3            & 88.1          \\
MetaCleaner              & 84.9            & 87.1            & 88.3     \\
CurriculumNet            & 85.2            & 86.8            & 87.9            \\
SOMNet                   & 86.0            & 87.9            & 88.6         \\
DivideMix                & 86.2            & 88.0            & 89.1     \\
LearnToLearn             & 85.3            & 87.6            & 88.4     \\
NLNL                     & 85.4            & 87.9            & 88.2     \\
JoCoR                    & 85.9            & 87.9            & 88.3   \\ \cdashline{1-4}[1.5pt/3pt]
MetaBaseline             & 85.8            & 87.1            & 88.7             \\
MAML                     & 84.6            & 86.8            & 87.9          \\
SNAIL                    & 84.1            & 86.3            & 87.2         \\
Distillation             & 83.7            & 85.9            & 87.3       \\
Finetune                 & 84.2            & 86.5            & 87.6      \\ \cdashline{1-4}[1.5pt/3pt]
Fusion1   & 86.8            & 88.3         & 89.7        \\
Fusion2   & 86.9            & 88.5         & 90.0        \\
\cdashline{1-4}[1.5pt/3pt]
SimTrans            & \textbf{89.8}         & \textbf{91.7}         & \textbf{92.8}   \\ \hline
\end{tabular}
\label{tab:SOTA}
\end{minipage}
\vspace{-15pt}
\end{table}

We compare with two types of methods: webly supervised learning and transfer learning.

For webly supervised learning, we compare with recent representative methods: SelfLearning \cite{SelfLearning}, MetaCleaner \cite{MIL3}, CurriculumNet \cite{CurriculumNet}, SOMNet \cite{SOMNet}, DivideMix \cite{DivideMix}, LearnToLearn \cite{LTL}, NLNL \cite{NLNL}, and JoCoR \cite{JoCoR}.
Webly supervised learning methods only use novel training set. 

For transfer learning across categories, there are mainly two groups of methods: few-shot learning and knowledge transfer.
% For few-shot learning, the ``shots'' in our setting are adequate but noisy, so we slightly adapt the few-shot learning baselines.
For few-shot learning, we compare with MetaBaseline \cite{MetaBaseline}, MAML \cite{FSL_optim_1}, and SNAIL \cite{SNAIL}.
MetaBaseline is a metric-based method, and we use the averaged feature of each novel category as class centroid following \cite{MetaBaseline}.
SNAIL is a memory-based method, and we use $16$ prototypes of each novel category as $16$ shots due to hardware limitation.
% SNAIL is a memory-based method and we have to select prototypes as shots for each novel category due to the limitation of memory size.
% Specifically, we select $16$ prototypes using the density-based method \cite{SelfLearning}.
% In addition, we pre-train the backbones for all few-shot learning baselines on novel training set, which is necessary for them to achieve competitive performances.
For knowledge transfer, the \textit{Distillation} baseline \cite{SAKE} enhances the classifier for novel categories by additionally predicting category distribution over base categories, which is supervised by a classifier pre-trained on base training set.
Another natural baseline, named as \textit{Finetune}, trains the model on the mixture of base training set and novel training set, and then fine-tunes on novel training set.

In addition, as far as we are concerned, there is no method which could jointly denoise web training data and transfer across categories without additional information (\emph{e.g.}, word vector). 
So we combine competitive methods in their own tracks using late fusion (\emph{i.e.}, average prediction scores).
The fusion ``SOMNet+MetaBaseline'' is referred to as \textit{Fusion1}, while the fusion ``DivideMix+MetaBaseline'' is referred to as \textit{Fusion2}.

% \subsubsection{Experimental Results}
All the results are summarized in Table \ref{tab:SOTA}. We also include the basic baseline \textit{Cls Loss}, which is trained on novel training set with standard classification loss (row 1 in Table~\ref{tab:ablation}).
Based on Table \ref{tab:SOTA}, we have the following observations: 

\noindent 1) Webly supervised learning methods outperform the basic baseline \textit{Cls Loss}, showing the general improvement brought by denoising web data without any support.

\noindent 2) Few-shot learning methods and knowledge transfer methods outperform the basic baseline \textit{Cls Loss}, indicating the advantage of transfer learning across categories.
Note that \textit{MetaBaseline} achieves a commendable performance, probably because  averaging features could denoise web images to some extent.
Nevertheless, the overall performances of transfer learning methods are limited by severe label noise.

\noindent 3) The simple combination of webly supervised learning and transfer learning, \emph{i.e.}, \textit{Fusion1} or \textit{Fusion2}, outperforms the above baselines, which directly demonstrates the effectiveness of our weak-shot setting. Furthermore, our method achieves the optimal performance against all baselines, which indicates the superiority of sample weighting and graph regularization based on similarity transfer.

\subsection{Qualitative Analysis for Weight and Graph}
We visualize to verify whether the higher weights are assigned to clean images and the transferred similarities could reveal semantic relations.
The visualizations and analyses are left to Appendix.

\subsection{Robustness Analysis of The Method}
We analyse the robustness of our method to several factors, including the number of web images, the selection of hyper-parameters, the noise ratio in web images, and the impact of different backbones.

For the impact of the number of web images, in our setting, the training data consists of clean images for base categories and web images for novel categories. 
The impact of the former is investigated in Section \ref{sec:base_scale}, and here we conduct experiments using different numbers of web images for each novel category on CUB dataset.
As shown in Table~\ref{tab:webnumber}, the general performances are gradually saturated given more training images (\emph{e.g.}, more than $800$ web images), and the observations of performance promotions against baselines are consistent across different numbers of web images.

For other factors (\emph{i.e.}, hyper-parameter, noise ratio, and backbones), we left the analysis to Appendix considering the space limitation.
% \subsection{Hyper-parameter Analysis}
% Our method mainly involves four hyper-parameters: $\alpha$ for balancing loss terms in Section~\ref{section:alpha}, $\beta$ for balancing loss terms in Section~\ref{section:simnet}, $C_m$ and $B$ for in mini-batch construction in Section~\ref{section:simnet}.
% The robustness analyses of our method \emph{w.r.t.} hyper-parameters are left to Appendix.

% \subsection{The Impact of the Noise Ratio in Web Images}
% To explore the impact of the noise ratio on our method, we conduct experiments on synthetic noisy data with different noise ratios, and the experiments and analyses can be found in Appendix.

% \subsection{The Impact of Different Backbones}
% We conduct experiments with different scales of backbones to show their impacts, and the detailed results and analyses are left to Appendix.

% \subsection{Extension to Generalized Setting}
% Recall that we have introduced the setting of generalized weak-shot learning in section \ref{section:definition}.
% % In this setting, we additionally include base test set (see Table~\ref{tab:datasets}) from base categories in the test set, which is more practical in real-world applications. 
% In this setting, the training (\emph{resp.}, test) set is the combination of base training (\emph{resp.}, test) set and novel training (\emph{resp.}, test) set (see Table~\ref{tab:datasets}), which is more practical in real-world applications. 
% We leave the results and analyses to Appendix.

\begin{table}[t]
\centering
\caption{Accuracies (\%) of different methods on three datasets in the generalized weak-shot setting. The best results are highlighted in boldface.
}
\begin{tabular}{c|ccc}
\hline \hline
Method                & {Car} & {CUB} & {FGVC} \\ \hline
Cls Loss   & 76.3         & 73.3         & 76.9            \\
Fusion1       &   77.5     &   74.8      &  78.8  \\
Fusion2 &  78.1        &    75.0      &  79.3      \\ \hline
SimTrans     &   \textbf{80.4}     &  \textbf{76.9}     & \textbf{81.7}  \\ \hline
\end{tabular}
\label{tab:Gen}
\vspace{-15pt}
\end{table}

\subsection{Extension to Generalized Setting} \label{sec:generalized}
In the generalized weak-shot learning, we additionally include base test set from base categories in the test set, so that test images may come from a mixture of base categories and novel categories, which is more practical in real-world applications.
We extend our method and baselines to generalized weak-shot learning scenario for comparison.

For simplicity, we directly conduct the experiments by treating the base training set as web images with low noise rate.
For our method, the first training stage of training SimNet remains the same, while the second training stage additionally includes base training set when training the main classifier. 
Thus, the sample weights assigned to base training images would be near one, and the graph regularization involving base training images will also do no harm to feature learning at least.

We compare with the basic baseline \textit{Cls Loss} as well as two combinations of webly supervised learning method and transfer learning method across categories.
% SOMNet and DivideMix are also conducted by treating base training images as web images, and their  clean probability thresholds for base training images are set as $0$.
SOMNet and DivideMix are also conducted by treating base training images as web images (the clean probability thresholds are set as $0$ for base training images in DivideMix).
MetaBaseline could be directly applied in the generalized setting.
One practical problem is that the image number of base categories and the image number of novel categories are highly imbalanced (\emph{e.g.}, $30$ and $1000$ for CUB), so we weigh the classification loss of each category using their image numbers (higher weight for the category with fewer images) for all methods in this setting.
The results are reported in Table \ref{tab:Gen}, from which we can find that the fusion of transfer learning and webly supervised learning outperforms the basic baseline, while our method further improves the results by a large margin.

\section{Conclusion}
In this paper, we consider the problem of learning novel categories from easily available web images with the support of a set of base categories with off-the-shelf well-labeled images.
Specifically, we have proposed to employ the pairwise semantic similarity to bridge the gap between base categories and novel categories.
Based on the transferred similarities, two simple yet effective methods
% , \emph{i.e.}, sample weighting and graph regularization, 
have been proposed to deal with label noise when learning the classifier on web data.
Adversarial loss is also applied to enhance the transferability of similarity.
Extensive experiments on four datasets have demonstrated the potential of our learning paradigm and the effectiveness of our method.

\section*{Acknowledgements}
The work was supported by the National Key R\&D Program of China (2018AAA0100704), National Natural Science Foundation of China (Grant No. 61902247), the Shanghai Municipal Science and Technology Major Project (Grant No. 2021SHZDZX0102) and the Shanghai Municipal Science and Technology Key Project (Grant No. 20511100300).

% \small{
% \bibliographystyle{plainnat}
% \bibliography{egbib_short}
% }
\small{
\bibliographystyle{plainnat}

}

\end{document}

\maketitle

In this appendix, we first formulate the optimization of adversarial similarity net in Section~\ref{sec:adv_simnet}. 
We clarify the dataset details and implementation details in Section~\ref{sec:dataset} and \ref{sec:implement}.
Then, we analyse the impact of hyper-parameters in Section~\ref{sec:hyperparameter}.
Afterwards, we report the results on ImageNet dataset in Section~\ref{sec:exp_imagenet}.
We provide qualitative analysis for the sample weights and similarity matrix learnt by our method in Section~\ref{sec:exp_qualitative}. 
Afterwards, we investigate the impact of different noise ratios in Section~\ref{sec:ratio} and explore the impact of different backbones in Section~\ref{sec:backbone}.
% Then, we conduct experiments for the generalized weak-shot learning in Section~\ref{sec:generalized}.
%After that, we describe the limitations of our work in Section~\ref{sec:limitation}.
%Finally, we show that our method is statistically significant in Section~\ref{sec:significance_test}.

\section{Adversarial Similarity Net} \label{sec:adv_simnet}
Here we introduce the optimization details of adversarial similarity net (SimNet) in Section $4.1$ of the main paper.
We denote the similarity between the $i$-th image $\textbf{x}_i$ and the $j$-th image $\textbf{x}_j$ as
\begin{equation}
    s_{i,j}=P(G(\textbf{x}_i,\textbf{x}_j))=P(\textbf{r}_{i,j}),
\end{equation}
where $G(\cdot)$ generates the relation feature $\textbf{r}_{i,j}=G(\textbf{x}_i,\textbf{x}_j)$ given an image pair, and $P(\cdot)$ outputs the similarity score $s_{i,j}$ given the relation feature.
The discriminator can be represented by
\begin{equation}
    d_{i,j}=D(\textbf{r}_{i,j}),
\end{equation}
where $d_{i,j}$ is the discriminator score indicating that the relation feature $\textbf{r}_{i,j}$ comes from base training set instead of novel training set.
To avoid confusion, we use the super-scripts $b$ and $n$ to distinguish the variables for base training set and novel training set.
In each training iteration, we construct mini-batch $X^b$ and $X^n$ with size $M$, and perform the following two optimizing steps in an alternating manner:

Firstly, we freeze the generator $G(\cdot)$ and update the discriminator $D(\cdot)$ by minimizing
\begin{equation}
\begin{aligned}
    L_D= &\frac{1}{M^2}\sum_{i=1}^{M}{\sum_{j=1}^{M}{\left[{\rm log} (1-D(\textbf{r}^b_{i,j}))+{\rm log} D(\textbf{r}^n_{i,j})\right]}}.
\end{aligned}
\end{equation}

Secondly, we freeze the discriminator $D(\cdot)$ and update the whole SimNet  by minimizing
% \begin{eqnarray}\label{eqn:L_G}
%     L_G=\frac{1}{M^2} \sum_{i=1}^{M}\sum_{j=1}^{M} \!\!\!\!\!\!\!\!\!\!&&[\beta {\rm log} D(\textbf{r}^b_{i,j})- c_{i,j}{\rm log} P(\textbf{r}^b_{i,j})-(1-c_{i,j}){\rm log} (1-P(\textbf{r}^b_{i,j})) ],
% \end{eqnarray}
\begin{eqnarray}\label{eqn:L_G}
    L_G=-\beta L_D+\frac{1}{M^2} \sum_{i=1}^{M}\sum_{j=1}^{M} \!\!\!\!\!\!\!\!\!\!&&[- c_{i,j}{\rm log} P(\textbf{r}^b_{i,j})-(1-c_{i,j}){\rm log} (1-P(\textbf{r}^b_{i,j}))],
\end{eqnarray}
where $c_{i,j}$ is the binary relation category label ($0$ for ``dissimilar pair" and $1$ for ``similar pair"),
and $\beta$ is a trade-off parameter set as $0.1$ via cross-validation.

\section{Datasets} \label{sec:dataset}
We evaluate our method on three popular fine-grained datasets. 1) CompCars \cite{Car} (Car for short): we regard the $431$ car models as fine-grained categories; 2) CUB \cite{CUB}: the standard $200$ bird species are regarded as fine-grained categories; 3) FGVC \cite{FGVC}: we regard the $100$ aircraft variants as fine-grained categories. In terms of base/novel category split, for CUB, we follow \cite{Split,Meets}, which is commonly used for zero/few-shot learning.
For Car and FGVC, we randomly split base and novel categories by $3:1$, with the same split ratio as in \cite{Split,Meets}.

Then, we need to construct web training set for novel categories. For Car, we use the released web images in WebCars \cite{MIL2}, where the noise ratio is about 30\% according to some sampled images from a few categories~\cite{MIL2}.
For CUB and FGVC, the datasets used in \cite{Unreasonable} only provide image URLs, most of which have expired, so we construct the web training images by ourselves.
In particular, we use the category names as queries to obtain $1000$ images from Google website after performing near-duplicate removal \cite{Duplicate,Meets}.

\begin{figure}[t]
\begin{center}
\includegraphics[width=1\linewidth]{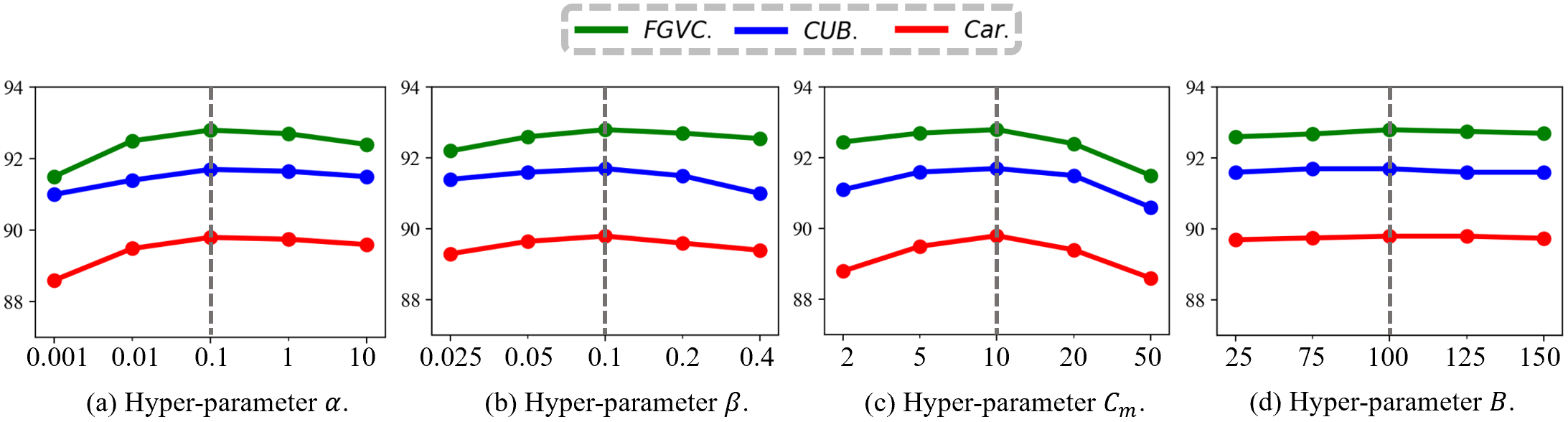}
\end{center}
\caption{The effects of varying the values of $\alpha$ (a) and $\beta$ (b) on three datasets. The dashed vertical lines denote the default values used in our paper.}
\label{fig:hyper}
\end{figure}

\section{Implementation Details} \label{sec:implement}
The proposed approach is implemented in Python $3.7$ and Pytorch $1.0.0$ \cite{Pytorch}, on Ubuntu $18.04$ with $32$ GB Intel $9700$K CPU and two NVIDIA $2080$ti GPUs.
Standard data augmentation is applied for all methods on all datasets, including random rotation, random resized crop, and random horizontal flip. 
Analogous to \cite{ClusterNet}, the backbone of similarity net is pre-trained on novel training set of the corresponding dataset.
We choose the classification accuracy as the evaluation metric for the main classifier, following \cite{Meets}.
We choose precision rate, recall rate, and F1-score as the evaluation metrics for similarity net, following \cite{ClusterNet}.

To select optimal hyper-parameters, we follow the validation strategy used in \cite{Meets,validation}. Assume that there are $C_b$ base categories and $C_n$ novel categories, we choose the first $C_v$ categories ($C_v=\lfloor \frac{C_n*C_b}{C_b+C_n} \rfloor$) based on the default category indices from $C_b$ base categories as validation categories. 
As a consequence, we need to further collect web images for validation categories.
In the validation stage, we regard $C_v$ categories as novel categories and $C_b-C_v$ categories as base categories. 
Then, we determine the optimal hyper-parameters according to the validation performance via random search \cite{Meets} within appropriate range.
For learning similarity net (SimNet) on base training set, we adopt the SGD optimizer with learning rate $0.01$ and batch size $100$ to train $50$ epochs for all datasets. 
For learning the main classifier on novel training set, we adopt the SGD optimizer with learning rate $0.005$ and batch size $128$. We train $50$ epochs for CUB and FGVC datasets, while training $100$ epochs for Car datasets.
We set weight decay as $0.0001$ and momentum as $0.9$ for both SimNet and the main classifier on all datasets.

\section{Hyper-parameter Analysis} \label{sec:hyperparameter}
In this section, we analyse the hyper-parameters: $\alpha$ for balancing graph regularization loss (see Eqn. (4) in the main paper), $\beta$ for balancing adversarial loss (see Eqn.~(\ref{eqn:L_G}) in Appendix), $C_m$ and $B$ in Section~4.1 of main paper for training SimNet.
The optimal values are determined via cross-validation as discussed in Section~\ref{sec:implement} and kept the same on all datasets (The batch sizes $B$ are with compatible learning rates).
As shown in Figure \ref{fig:hyper}, we vary each hyper-parameter in a range while fixing the other hyper-parameters, and plot the results obtained by our full-fledged method.
The results reported in Figure \ref{fig:hyper} suggest that the performance of our method is robust to the hyper-parameters in an appropriate range.

\section{Experiments on ImageNet dataset} \label{sec:exp_imagenet}
Note that the focus of this work is fine-grained classification, so we mainly conduct experiments on fine-grained datasets. 
In this section, we explore the effectiveness of our setting and method on a large-scale dataset ImageNet \cite{ImageNet}, even though it is not fine-grained.

Following the random split in \cite{ImageNetSplit,ClusterNet}, we have $882$ categories and $118$ categories for base set and novel set, respectively.
The base training/test set and the novel test set are from the ImageNet dataset \cite{ImageNet} while the novel training set is constructed using WebVision $1.0$ \cite{WebVision} (about 20\% noise ratio according to \cite{WebVision}).
Specifically, each base category contains $1279$ clean training images and $50$ test images on average, while each novel category  contains $2423$ web training images and $50$ test images on average.

We compare with the basic baseline \textit{Cls Loss} as well as two combinations of webly supervised learning method and transfer learning method across categories, \emph{i.e.}, \textit{SOMNet+MetaBaseline} and \textit{DivideMix+MetaBaseline}, considering their competitive performance reported in the main paper.
The results are listed in Table \ref{tab:ImageNet}, from which we can find that our method still dramatically improves the results, even on a coarse-grained dataset.

\begin{table}[t]
\centering
\caption{Accuracies of different methods on ImageNet dataset in the weak-shot setting. The best results are highlighted in boldface.
}
\begin{tabular}{c|c}
\hline \hline
Method                & Accuracy (\%) \\ \hline
Cls Loss   & 74.2             \\
SOMNet+MetaBaseline       &   75.1 \\
DivideMix+MetaBaseline &  76.5       \\ \hline
SimTrans     &   \textbf{78.7} \\ \hline
\end{tabular}
\label{tab:ImageNet}
\end{table}

\section{Qualitative Analysis for Sample Weight and Similarity Matrix} \label{sec:exp_qualitative}
\begin{figure}[t]
\centering
\begin{center}
\includegraphics[width=0.6\linewidth]{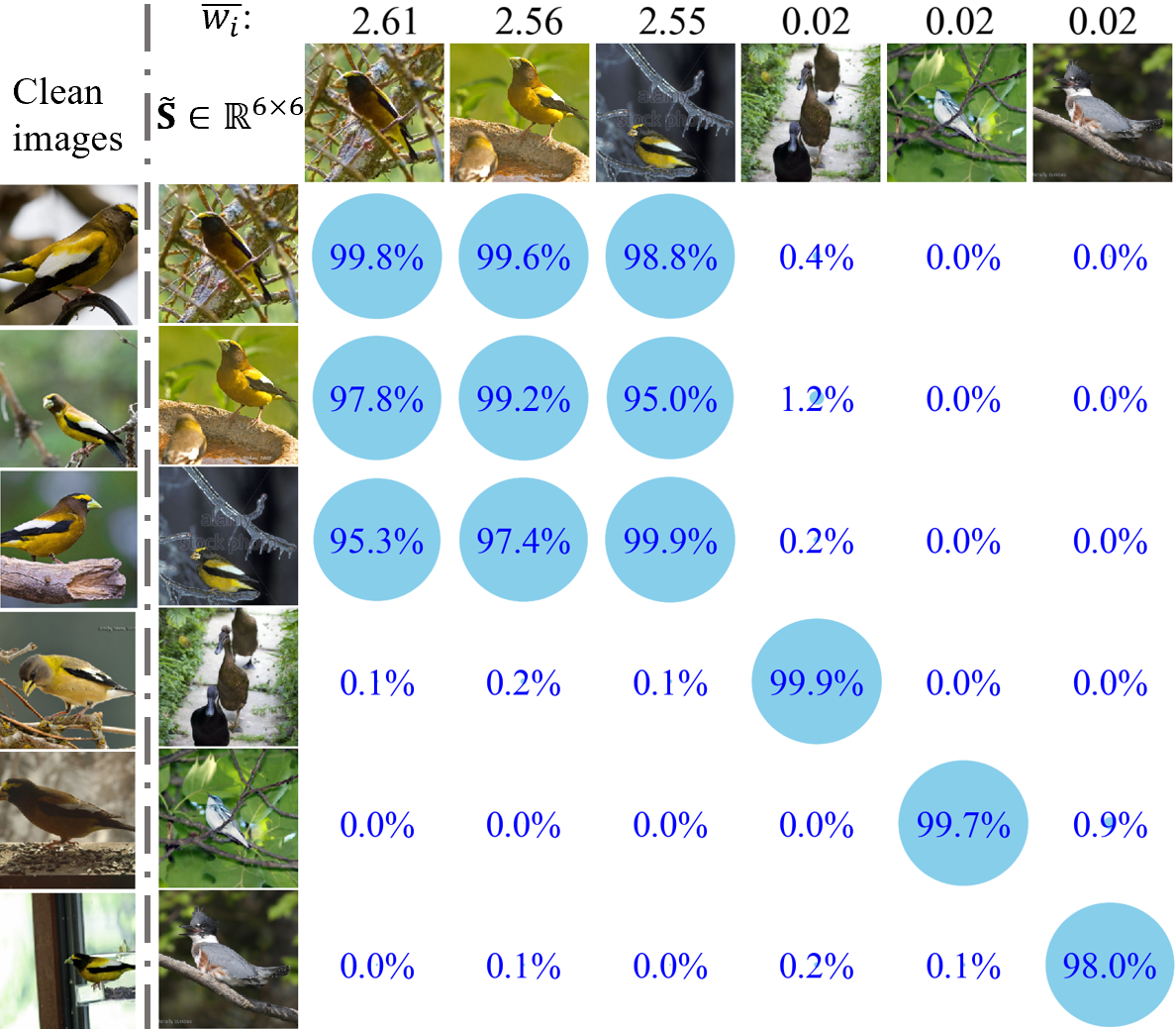}
\end{center}
\caption{Visualization of sample weights and similarity matrix of $6$ web images ($3$ with the highest weights and $3$ with the lowest weights) from the category ``Evening Grosbeak''. The sample weights are shown on top of the web images. The right part shows the $6\times 6$ similarity matrix.
The left column shows some clean test images of the visualized category for reference.
}
\label{fig:WG}
\end{figure}

According to our hypothesis and formulation (Eqn. (1) in the main paper), in our weighted classification loss, higher weights are more likely to be assigned to non-outliers. Besides, the transferred similarities could reveal the semantic relations among web images.
For qualitative analysis of assigned weights, by taking the ``Evening Grosbeak'' category in CUB as an example, we rank all web training images by assigned weights.
According to the rank, we show the first $3$ and the last $3$ images as well as their similarity matrix in Figure \ref{fig:WG}.
On the one hand, we could observe that the images with high weights are very similar to clean images, while the images with low weights are outliers.
On the other hand, we observe that the transferred similarities accurately portray the semantic relations among web images.
We have similar observations for the other categories and on the other datasets.

\begin{table}[t]
\centering
\caption{Accuracies (\%) on various levels of noise. The best results are highlighted in boldface.}
\begin{tabular}{c|cccc}
\hline \hline
Noise Ratio            & 10\% & 20\% & 30\% & 40\% \\ \hline
Cls Loss               & 77.1    &  74.7    & 71.1     &  66.0    \\
SOMNet+MetaBaseline          & 78.0    &  76.0    & 73.8     &  69.4    \\
DivideMix+MetaBaseline       & 78.2    &  76.4    & 74.3     &  70.2    \\ \hline
SimTrans        & \textbf{79.4}    &  \textbf{78.3}    & \textbf{76.4}     &  \textbf{73.6}    \\ \hline
\end{tabular}
\label{tab:noiselevel}
\end{table}

\section{Experiments with Different Noise Ratios} \label{sec:ratio}
In practice, we have no ground-truth labels for web data and expert knowledge is especially required to annotate or check labels for fine-grained categories, so it is hard to correct the labels and change the noise ratio of fine-grained web datasets.
Therefore, to control the noise ratios, we can only conduct experiments on synthetic noisy data instead of real-world web data \cite{Meets,Unreasonable}. 
Following \cite{NLNL}, we synthesize noisy training data with different noise ratios using CIFAR-100.
Specifically, as in \cite{ImageNetSplit,ClusterNet}, we randomly split the classes into base set and novel set with the same split ratio stated in Section~\ref{sec:dataset}.
For label noise, we circularly flip each class into the next within each super-class of CIFAR-100, following \cite{NLNL}.
We summarize the results in Tab.~\ref{tab:noiselevel}, where we could see that our method is generally effective for different noise ratios.

\section{Impact of Different Backbones.}  \label{sec:backbone}
We conduct experiments using different backbones on CUB dataset, including ResNet18, ResNet34, ResNet50, and ResNet101.
We compare with the two most representative and competitive baselines, \emph{i.e.}, ``SOMNet+MetaBaseline'' and ``DivideMix+MetaBaseline''.
Are reported in Tab. \ref{tab:Backbone}, our method could achieve general improvement for different scales of architectures.

\begin{table}[t]
\centering
\caption{Accuracies (\%) of different methods on four scales of backbone on CUB dataset. The best results are highlighted in boldface.
}
\begin{tabular}{c|cccc}
\hline \hline
Method                & ResNet18 & ResNet34 & ResNet50 & ResNet101 \\ \hline
Cls Loss                & 80.7  & 83.3  & 85.4 & 85.9 \\
SOMNet+MetaBaseline     & 82.8  & 85.9  & 88.3 & 89.2 \\
DivideMix+MetaBaseline  & 83.6  & 86.4  & 88.5 & 89.3 \\ \hline
SimTrans & \textbf{88.3} & \textbf{90.4} & \textbf{91.7} & \textbf{92.1} \\ \hline
\end{tabular}
\label{tab:Backbone}
\end{table}

\bibliographystyle{plainnat}

% }